\documentclass[sigconf]{acmart}
\setcopyright{none}
\renewcommand\footnotetextcopyrightpermission[1]{} 

\usepackage{amsmath,amsfonts, pifont}
\usepackage{graphicx}
\usepackage{textcomp}
\usepackage{xcolor}
\usepackage{algorithm}
\usepackage{algpseudocode}
\def\algbackskip{\hskip-\ALG@thistlm}
\makeatother
\usepackage{caption}
\usepackage{subcaption}
\usepackage{booktabs}
\usepackage{makecell}

\DeclareMathOperator*{\argmaxA}{argmax}
\DeclareMathOperator*{\maxA}{max}

\newcommand{\cmark}{\ding{52}}
\newcommand{\xmark}{\ding{55}}

\def\BibTeX{{\rm B\kern-.05em{\sc i\kern-.025em b}\kern-.08em
    T\kern-.1667em\lower.7ex\hbox{E}\kern-.125emX}}

\AtBeginDocument{%
  \providecommand\BibTeX{{%
    \normalfont B\kern-0.5em{\scshape i\kern-0.25em b}\kern-0.8em\TeX}}}

\copyrightyear{2023} 
\acmYear{2023} 
\setcopyright{rightsretained} 
\acmConference[CHASE '23]{ACM/IEEE International Conference on Connected Health: Applications, Systems and Engineering Technologies}{June 21--23, 2023}{Orlando, FL, USA}
\acmBooktitle{ACM/IEEE International Conference on Connected Health: Applications, Systems and Engineering Technologies (CHASE '23), June 21--23, 2023, Orlando, FL, USA}
\acmDOI{10.1145/3580252.3586979}
\acmISBN{979-8-4007-0102-3/23/06}

\begin{document}

\title{Active Reinforcement Learning for Personalized Stress Monitoring in Everyday Settings}

\author{Ali Tazarv}
\email{atazarv@uci.edu}
\orcid{0001-7176-791X}
\affiliation{%
  \institution{University of California Irvine}
  \state{California}
  \country{USA}
}

\author{Sina Labbaf}
\email{slabbaf@uci.edu}
\affiliation{%
  \institution{University of California Irvine}
  \state{California}
  \country{USA}}

\author{Amir Rahmani}
\email{a.rahmani@uci.edu}
\affiliation{%
  \institution{University of California Irvine}
  \state{California}
  \country{USA}}

\author{Nikil Dutt}
\email{dutt@ics.uci.edu}
\affiliation{%
  \institution{University of California Irvine}
  \state{California}
  \country{USA}}

\author{Marco Levorato}
\email{levorato@uci.edu}
\affiliation{%
  \institution{University of California Irvine}
  \state{California}
  \country{USA}}


\begin{abstract}
  Most existing sensor-based monitoring frameworks presume that a large available labeled dataset is processed to train accurate detection models. However, in settings where personalization is necessary at deployment time to fine-tune the model, a person-specific dataset needs to be collected online by interacting with the users. Optimizing the collection of labels in such phase is instrumental to impose a tolerable burden on the users while maximizing personal improvement. In this paper, we consider a fine-grain stress detection problem based on wearable sensors targeting everyday settings, and propose a novel context-aware active learning strategy capable of jointly maximizing the meaningfulness of the signal samples we request the user to label and the response rate. We develop a multilayered sensor-edge-cloud platform to periodically capture physiological signals and process them in real-time, as well as to collect labels and retrain the detection model. We collect a large dataset and show that the context-aware active learning technique we propose achieves a desirable detection performance using $88\%$ and $32\%$ fewer queries from users compared to a randomized strategy and a traditional active learning strategy, respectively.
\end{abstract}




\maketitle
\pagestyle{plain} 

\section{Introduction}
\label{sec: intro.}
Assessment of physiological stress is one of the first steps to monitoring individuals' well-being and determining effective interventions and recommendations to promote mental and physical health. Traditionally, such assessments have been performed through self-reported surveys \cite{yu2022semi}. However, recent developments in wearable technologies and bio-sensors, along with the characterization of the correlation of these bio-signals with physiological stress, have opened new opportunities to devise practical systems and models that automatically detect stress. There is an extensive body of research on the detection and prediction of mental well-being based on bio-signals both in controlled laboratory settings with an external stimulus \cite{sarkar2021detection, kalra2020mental, schmidt2018introducing}, and in normal everyday settings \cite{tazarv2021personalized, wu2021multi, yu2020passive, han2020objective, umematsu2020forecasting, bogomolov2014daily, yu2022semi, mundnich2020tiles, sah2022adarp}. However, the latter is a highly challenging task, mainly due to the barriers in data collection and labeling the collected data, as demonstrated later in this paper. 
In fact, in everyday settings, biosignals have distinct distributions among individuals due to the differences in physiology and lifestyle, as well as across the multitude of activities an individual may engage with. Moreover, individuals comprehend stress levels differently, which results in a bias in the collected labels. 
For these reasons, there is a strong demand to \emph{personalize} predictive models -- to the specific person through the broad range of activities the individual engages in -- as we show in Section~\ref{sec: motivation}. However, since such considerations are necessarily performed at deployment time, conceptual and technical challenges arise.

Given that self-reported stress (ecological momentary assessment (EMA)) is considered the gold standard, the source of the information needed for personalization is the user. Therefore, the user requires to respond to real-time queries associating the data collected by the sensors to stress labels in everyday settings. These necessities require the system  to support collecting biosignals and contextual parameters, triggering and collecting EMAs, and retraining and updating the model. However, user adherence is a difficult component in such systems. In fact, the monitored person may fail to respond to frequent labeling requests, which may also increase the chance of a drop-out. Moreover, as the individuals engage in their daily activities, it is likely that the response to EMAs may go unanswered, or is completed with delay, thus reducing the availability of labels, and their correlation with the biosignals.

This work aims to address the above problems by leveraging active learning (AL). Traditional AL refers to the online problem of selecting the most meaningful samples in a dataset to minimize learning time, assuming the full dataset is available. Typically, sample selection methodologies focus on ``hard'' samples close to the decision boundary of the current classifier version to increase the detection resolution in uncertain regions of the feature space. In our setting, the system needs to decide in real-time whether or not to request the label (i.e., triggering the EMA) to the monitored person, thus making decisions on a sample-by-sample basis.

Different from typical AL problems, in everyday settings, the user plays a crucial role in determining the sample-label pairs that are acquired,  as well as the label-to-sample correlation. In fact, the monitored subject may decide to ignore a request, or submit the response to the query with a large delay, thus reducing the correlation between the target sample and the label. To address this issue, we propose a \emph{contextual} form of active learning, where the system uses real-time data to factor in the decision-making about query submissions. 
In summary, the contributions of this paper are as follows:

\vspace{1mm}
\noindent
{\bf Infrastructure:} We develop a multi-layer infrastructure for the collection and labeling of physiological data from wearable sensors, as well as the real-time detection of stress. To enable the execution of machine learning models, and their retraining, we adopt a sensor-edge-cloud architecture, where the data locally collected by the sensors are relayed by the edge (a smartphone in our system) to the cloud for processing (detection, decision making and retraining). We characterize and describe in depth the capabilities and performance of the infrastructure.

\vspace{1mm}
\noindent
{\bf Dataset:} We conducted an IRB-approved human subject study and collected a total of more than 2,629 days worth of data in everyday settings from a group of college students.
The dataset includes Photoplethysmography (PPG) and motion data (acceleration, gyroscope, gravity), and it is partially labeled with stress level, emotions, and physical activity through EMAs at semi-random times. 

\vspace{1mm}
\noindent
{\bf Contextual Deep Q-Learning Based AL:} We propose a new form of AL, based on Deep Q-Learning, to incorporate the notion of effective interaction with the monitored person in the data collection phase of personalization. The ultimate objective of the agent is maximizing the performance of stress detection on the user. To this purpose, in the optimization objective we not only incorporate a measure of uncertainty from the current detector, but also the characteristics -- learned from data -- of the expected response from the user. The latter are obtained through processing contextual features of the samples in the dataset. Our results show that our contextual active learning agent reduces the number of required EMAs by up to $88\%$ compared to a random selection method, and up to $32\%$ compared to a traditional active learning method.

The rest of this paper is as follows: Section \ref{sec: motivation} includes a brief review of stress assessment and related work in the field, and a discussion on the need for personalizing these models and the importance of user behavior and contextual information in real-time label collection in everyday settings. Section \ref{sec: system_design} describes the platform we developed for data collection and processing. Section \ref{sec: caal_method} presents the proposed context-aware active learning method which can capture user behavior and contextual information in its query engine, as well as temporal correlation in the data, to optimize the query schedules. Section \ref{sec: dataset} describes the collected dataset. Section \ref{sec: results} presents the results of different query methods, on personalization. Section \ref{sec: conclusion} concludes the paper. 

\section{Motivation and Related Work}
\label{sec: motivation}

\vspace{1mm}
\noindent
{\bf Related Work:}
The term stress may be used in reference to an external (e.g. way of living, relationship problems, financial problems, etc.) or internal (e.g. personality structure, way of thinking, etc.) affairs that triggers negative emotions (worry, fear, etc.) and associated physiological changes. A physiological \textit{stress response} refers to the bodily changes elicited by environmental events or conditions, known as \textit{stressors}.
The multifaceted nature of stress can be decomposed into three main components: the \textit{psychological}, the \textit{behavioural} and the \textit{physiological}.
Traditionally, stress is measured using self reported surveys. For example, S. Cohen et~al. developed the Perceived Stress Scale (PSS) for measuring stress over the past month \cite{cohen1994perceived}. Holms and Rahe introduced a scale that estimates the total amount of stress in the past year based on self reported surveys \cite{holmes1967social}.

Survey based stress assessment methods are influenced by a multitude of systematic measurement errors such as the response bias (the tendency to respond in a manner influenced by the question). Besides, even though some behavioural bodily patterns (such as facial expressions) are manifested in response to stress, they may also be subject to intentional or even partially conscious control. Consequently, related recordings may also contain systematic errors when used to estimate the magnitude of the stress response. 

These limitations, along with the growing industry of high quality sensors technology enhance the need for reliable physiological stress detectors. Bio-signals features of stress related events are mostly involuntary; such measures can be sought through electrocardiography (ECG), Photoplethysmography (PPG), electromyography (EMG), skin conductance (SC) or electrodermal activity (EDA), respiratory (RSP), skin temperature (ST), pupil diameter (PD), and brain activity recorded through electroencephalographic(EEG) \cite{giannakakis2019review}.

Automatic assessment of physiological stress is carried out in two general settings: controlled settings \cite{kalra2020mental, schmidt2018introducing},  and everyday settings \cite{tazarv2021personalized, yu2020passive, han2020objective, umematsu2020forecasting, bogomolov2014daily, yu2022semi, mundnich2020tiles, sah2022adarp}. In controlled settings, there are certain situations and a limited time in which the biosignals are collected while the subject goes through various physiologically stressful events. Conversely, in everyday settings the subjects behave freely and do not perform a set of planned activities and events. Intuitively, the latter scenario is much more challenging from an automated detection perspective due to the broad range of activities the individual can engage in, and their implications on biosignals' characteristics.

The study of biosignals in everyday settings has gained popularity in recent years due to the improved availability of off-the-shelf bio-sensors for continuous use in daily life, which opened the opportunity for extracting useful information about mental and physical health.
For instance, Yu and Sano collect and study biosignals using a wristband wearable sensor in addition to data about smartphone usage from a large group of college students over periods of 30-90 days \cite{yu2020passive}. The surveys are collected once a day, and ask about participants' well-being and mental status. These data in addition to local weather data, are used to train a deep neural network and predict the well-being in the future and for new users. 
In another study~\cite{yu2022semi}, the authors collected data from a group of healthy participants for 5-9 days. The data are captured using a wrist-worn sensor and a chest patch, and short stress evaluation surveys are filled at predetermined times throughout the day. The authors then develop a semi-supervised learning method leveraging unlabeled data in addition to the labeled for stress detection. 
Mundnich, \textit{et~al.} have collected a dataset including bio-signals from wearable devices, behavioral information (from smartphones) and environmental information such as audio features from a group of hospital workers for 10 weeks \cite{mundnich2020tiles}. In daily surveys, participants are asked about their job, health, and psychological flexibility. 
In \cite{sah2022adarp}, Sah, \textit{et~al.} collected a total of 1698 hours of physiological sensor data from a population of participants with alcohol use disorders. The participants were asked to complete four daily surveys at prescheduled times about emotions, alcohol related cravings, and experiences of pain and discomfort. The authors then developed a deep neural network model to classify stressful vs non-stressful instances. 
Wang, \textit{et~al.} collected smartphone data for 12 months \cite{wang2020social}. The data include physical activity, location, ambient sound features and light intensity, as well as phone usage patterns, completed by self-reported surveys about their social functioning. 
Battalio, \textit{et~al.} \cite{battalio2021sense2stop} collect physiological and motion data and analyze it in real-time on the edge layer (smartphone) for 15 days and the surveys are collected at random times, after a certain number of self-reports of stress, and at the end of each day.

\vspace{1mm}
\noindent
{\bf Motivation:} 
Most of these contributions and methods collect and store the data, and independently request the completion of surveys to the participants at prescheduled or random times. Then, they process and analyze the data in an offline fashion. Importantly, the data and \emph{labels} refer to different time indexes and temporal time scales, where the labels (surveys) are typically at a very coarse granularity. Thus, those approaches are appropriate to relate biophysical signals to a general coarse-grain state of the person.

In this study, our objective is to build an effective infrastructure and detection methodology to infer episodic stress at a fine temporal granularity in an everyday setting. This is a task that presents several challenges. First, the data collection, communication and computing infrastructure becomes real-time and all its components are intertwined. As explained later, our method requires the infrastructure to be able to process the samples and labels as they are collected not only to support real-time detection, but also to optimize and control the interaction with the user. We detail the structure and performance of the multi-layered infrastructure we developed in the next section. 

\begin{figure}
    \centering
    \includegraphics[width=1\linewidth]{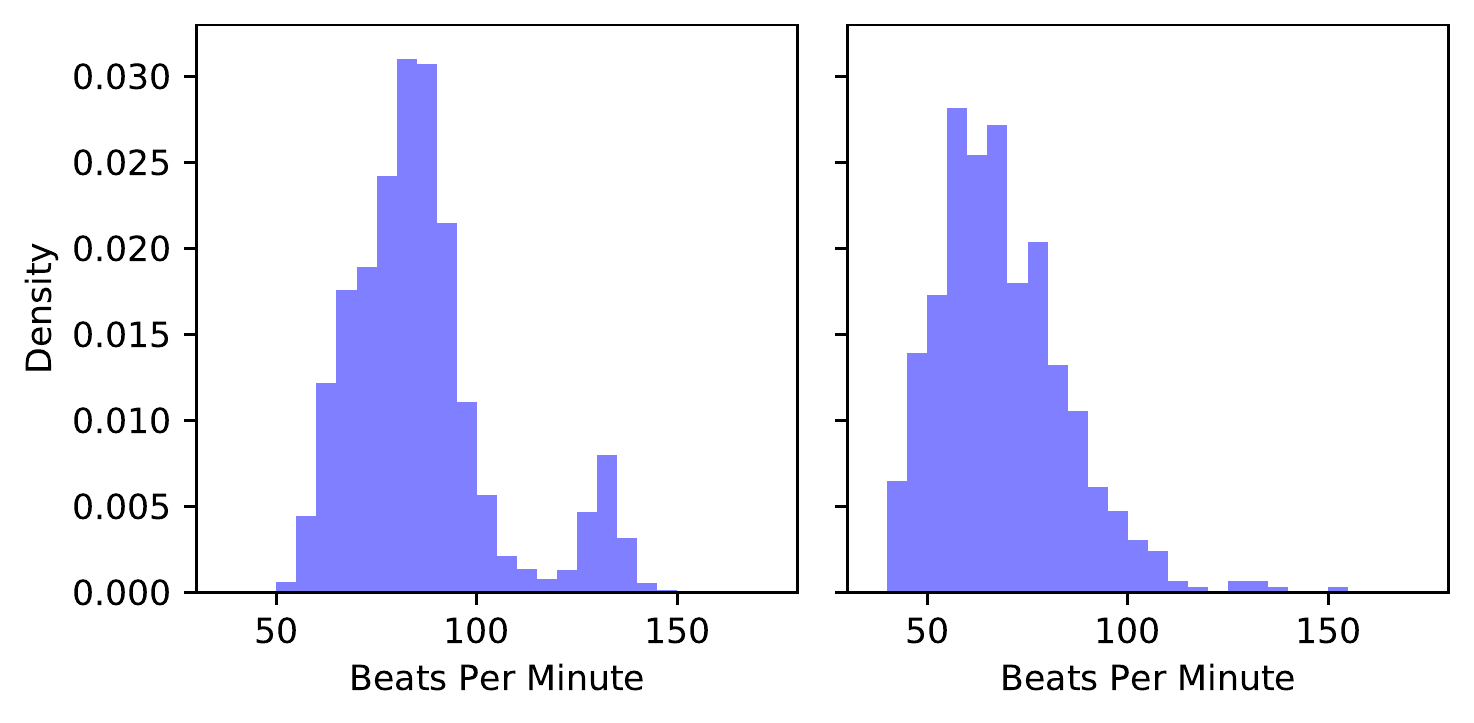}
    \caption{BPM for two subjects.}
    \label{fig: bpm}
\end{figure}

Furthermore, our study demonstrates that effective stress detection in this setting and at this time scale requires personalization. We illustrate this need here by presenting a preliminary analysis of the dataset we collected. Due to physiological and lifestyle differences, biosignals have different distributions among individuals. Fig.~\ref{fig: bpm} shows the distribution of \textit{beats-per-minute} for two individuals in duration of 90 days from our study, in which not only the range of hear-beat is different, but also the frequency pattern is significantly different in certain regions. At an intuitive level, such perceivable difference may complicate the task of a detector trained over data from a multitude of subjects without an emphasis on the specific target individual.

In addition, since labels are collected through surveys, the difference in response bias and living conditions among individuals results in a specific distribution of labels for each person, as shown in Table \ref{tab: labels_dist}. The difference in distributions of both signal and labels for each subject, results in a potential inaccuracy of the stress prediction models and personalizing an automatic well-being predictor using subjective data significantly improves the accuracy \cite{siirtola2020comparison, tazarv2021personalized}. 
\begin{table}[tbp]
    \centering
    \caption{Distribution of stress labels from different subjects.}
    \begin{tabular}{r|c c c c c c}
        Subject & S1 & S2 & S3 & S4 & S5 & S6\\ \hline
        not stressed & 156 & 348 & 95 & 136 & 99 & 942\\ \hline
        stressed & 4 & 28 & 25 & 4 & 20 & 68\\ \hline
        ratio of & $2.5\%$ & $7.4\%$ & $20.8\%$ & $2.9\%$ & $16.8\%$ & $6.7\%$ \\
        \textit{stressed} labels &&&&&&
    \end{tabular}
    \label{tab: labels_dist}
\vspace{-0.5cm}
\end{table}


In model personalization, we have a generalized model that is trained on data from multiple individuals, excluding the target subject \textit{A}, whose data are assumed unavailable at the initial deployment stage. As the data from this subject are added to the training set and the model is retrained, the model performance improves when tested on the same subject. Table~\ref{tab:personintro} shows the effect of personalization on recall (true positive rate) -- a key classification metric for imbalanced datasets -- as the number of labels available from the specific person grows. Recall increases from 0.23 to 0.44 when 600 labels from the subject are used. Note that recall here is measured on the positive (\textit{stressed}) class only. Clearly, detecting the majority class -- normal state in our usecase -- is a simpler task for the classifier. The dataset and detection model are described in detail later in the paper.

However, intuitively the main challenge is that collecting a large number of labels from a person would either require a large amount of time, or an excessive number of requests to the user -- with a clear impact on the usability of the system and rejection rate. Fortunately, when collecting the data from the specific subject, some instances have higher impact on improving the model compared to others. In offline settings, the identification of the useful portion of training to reduce training complexity is called Active Learning. Herein, we use AL to accelerate personalization while minimizing the number of queries sent to the user. We show that a fundamental aspect of our solution is the interaction with the user, that needs to be incorporated in the metrics used to drive the labeling requests.


\begin{table}[tbp]
    \centering
    \caption{Personalization for different number of subjective instances used. The metric is recall on the minority class.}
    \begin{tabular}{p{1.7cm}|c c c c c}
         No. instances &  0 (pre-trained) & 150 & 300 & 450 & 600 \\ \hline
         recall on minority class & 0.23 & 0.29 & 0.34 & 0.40 & 0.44\\
    \end{tabular}
    \label{tab:personintro}
\end{table}

We propose a real time data analysis platform that not only enables us to identify the useful instances, but also analyse contextual information from the subject and their response behavior in that context. Therefore, we are able to query only for instances that are expected to have a high impact on personalization and have a high probability to be fulfilled by the user based on captured  temporal patterns.

Table~\ref{tab: related_work} summarizes the studies mentioned earlier vs our study. \textbf{Data-based Queries} means the agent decides to query for each sample or not, after it analyzes the sample. \textbf{Personalized Queries} means the agent acts differently for each individual, based on their personal history and traits. \textbf{Online Model Update} refers to whether the prediction model is being personalized throughout the study or not. And \textbf{Personalized Detection} is also about prediction model. All of these experiments are performed in everyday settings, also known as \textit{in the wild}. 

\begin{table*}[tbp]
    \caption{Comparison of our study vs other studies}
    \begin{center}
        \begin{tabular}{|p{1.7cm}|p{2.8cm}|p{2.9cm}|p{1.3cm}|p{1.2cm}|p{1.75cm}|p{1.3cm}|p{1.75cm}|}
        \hline
             \textbf{Researchers} & \textbf{Data} & \textbf{EMA/Survey Schedule or Frequency}& \textbf{Realtime Analysis} & \textbf{Data-based Queries} & \textbf{Personalized Queries} & \textbf{Online Model Update} & \textbf{Personalized Detection} \\ \hline
             Yu, Sano \cite{yu2022semi} & ECG, SC, ST, Motion (acceleration) & prescheduled times, 1.5 hours apart, 10/day & \xmark & \xmark & \xmark & N/A & \xmark \\ \hline
             Yu, Sano \cite{yu2020passive} & SC, ST, Motion (acceleration), Mobile Phone data, weather & prescheduled time, 1/day (evening) & \xmark & \xmark & \xmark & N/A & \cmark \\ \hline
             Mundnich, et~al. \cite{mundnich2020tiles} & ECG, Motion (acceleration and gyroscope), Audio features, Light, Mobile Phone data & prescheduled time, 1/day & \xmark & \xmark & \xmark & N/A & N/A \\ \hline
             Sah, et~al. \cite{sah2022adarp} & PPG, SC or EDA, Motion (acceleration), ST & prescheduled times, 4/day, plus voluntary reports & \xmark & \xmark & \xmark & N/A  & \xmark \\ \hline
             Wang, et~al. \cite{wang2020social} & smartphone data including activity, location, sound level, phone usage, etc. & every 3 months & \xmark & \xmark & \xmark &  N/A & \xmark \\ \hline
             Battalio, et~al. \cite{battalio2021sense2stop} & ECG, Motion (acceleration, gyroscope), respiration & event triggered and random times and end of day, varies & \cmark & \cmark & \makecell{\\ start and end \\ of day only} & \xmark & \xmark \\ \hline
             \textbf{Ours} & \textbf{PPG, Motion (acceleration, gyroscope, gravity)} & \textbf{not fixed times, depends on the data, on average 5-6/day} & \cmark & \cmark & \cmark & \cmark & \cmark \\ \hline
        \end{tabular}
        \label{tab: related_work}
    \end{center}
\end{table*}

In this paper, since we particularly focus on physiological stress in everyday settings, feasibility of wearing the sensor in daily life for long periods of time is an important factor which needs to be considered. Among the sensor technologies mentioned earlier, PPG sensors provide better feasibility for the user, since they already exist in most wearable devices such as smart watches and smart rings. Therefore, in this study we only focus on estimation of physiological stress in everyday settings, using PPG signals.

\section{System Design and Challenges}
\label{sec: system_design}

The overall system we consider is composed of multiple layers: a \emph{sensor} layer to collect the biosignals; an \emph{edge} layer to relay the sensor data and interact with the user; and a \emph{cloud} layer to process the data, and perform detection and decision making. In our deployment, these layers respectively correspond to a smartwatch, a smartphone and a server.

\begin{figure}
    \centering
    \includegraphics[width=\linewidth]{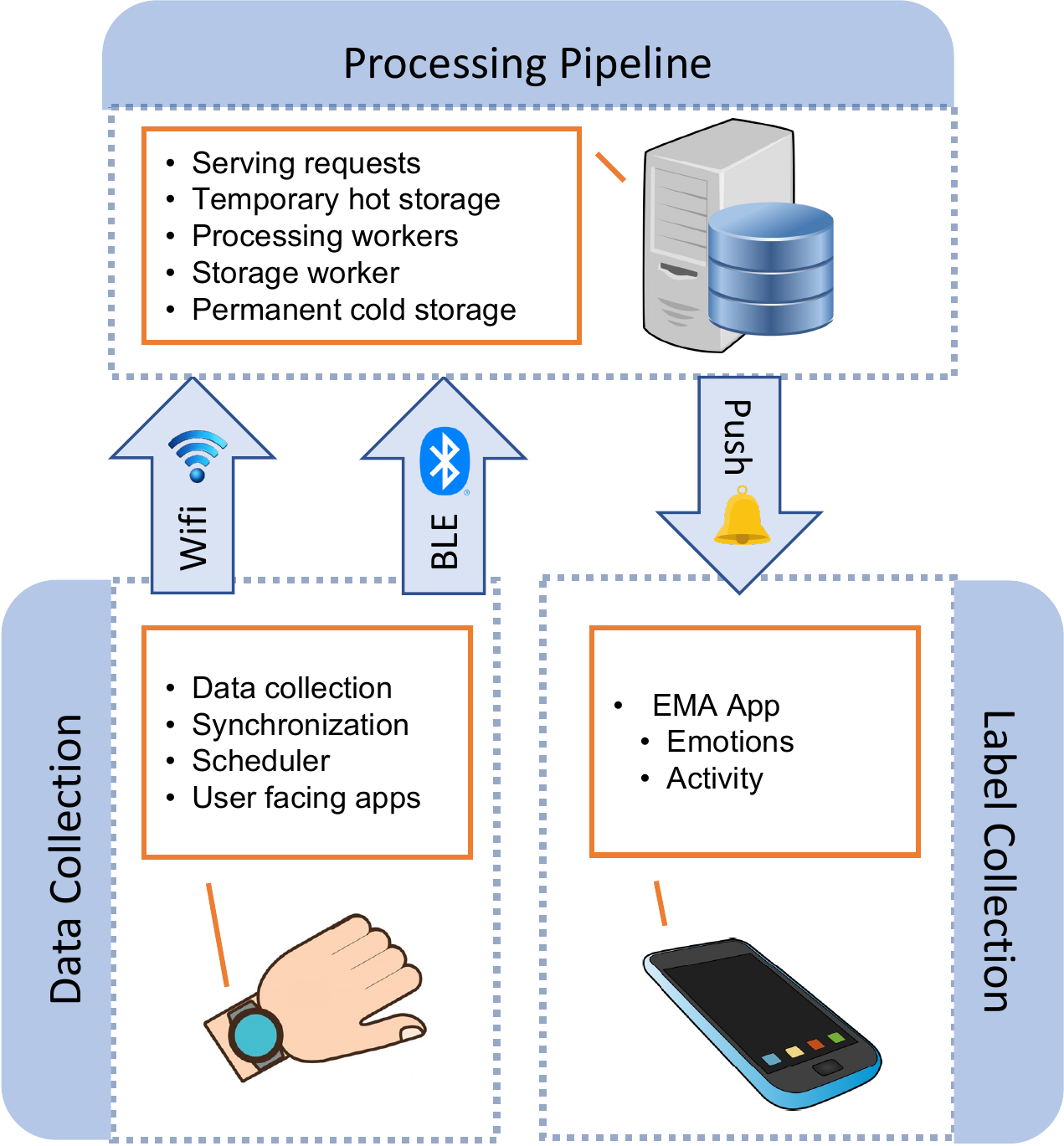}
    \caption{System overview}
    \label{fig: system}
\end{figure}

As discussed earlier, the task at hand requires the design of a system that forms a fine-grain close-loop with the monitored subjects. This system must incorporate three main components: data collection, label collection, and analysis and learning. An overview of the core of the system is provided in Fig.~\ref{fig: system}. Other components of the system that make the study feasible, but that are not essential to the control loop include the administrative dashboard and administrative API, which facilitate recruitment and coordination. In this article we will only focus on the core components. 

Our design focuses on three objectives: (\emph{i})
ability to collect the incoming data and promptly deliver them to the analysis unit; (\emph{ii}) Ability to process the data at scale; (\emph{iii}) Short response time to deliver queries.
In the following subsections we introduce these components and discuss our implementation principles.

\subsection{Data collection}
In this study, for data collection we used a Samsung Galaxy Gear Sport smart watch, a commercial smart watch suitable for longer studies compared to research-specific devices.
In order to collect raw data from the watch, we designed a collection of services and applications:
(\emph{1}) a service for the smartwatch that collects chunks of data from sensors. (\emph{2}) a service to synchronize data. 
(\emph{3}) A scheduling service that invokes the above two services at predetermine frequencies. (\emph{4}) An application to reset the scheduling service manually. (\emph{5}) An application to synchronize all the remaining data manually. We used Samsung's Tizen studio to develop all the software for the smartwatch.

\subsubsection{Signals}
We collect the following data from the following sensors:

\vspace{1mm}
\noindent
\emph{Photoplethysmography} (PPG): 
This sensor provides the value of reflection of light from the surface of the skin. This signal was collected at the maximum frequency supported by the smartwatch, that is, 20Hz.

\vspace{1mm}
\noindent
\emph{Accelerometer} (ACC):
This sensor returns three values of acceleration along three different dimensions.
This signal is also measured at 20Hz.

\vspace{1mm}
\noindent
\emph{Gyroscope} (GYR):
Gyroscope measures the angular velocity along the three axes.
We set the capture rate to 20Hz for this sensor as well.

\vspace{1mm}
\noindent
\emph{Pressure}:
The pressure sensor measures the atmospheric pressure at 10Hz.
\vspace{2mm}

We set our data collection service to be active for 2 continuous minutes and set the scheduler service to call this service every 15 minutes. 
This setting alongside other normal use makes the battery last for about two days on average.  We note that the battery lasts for about 3.5 days under normal smartwatch usage. 
Our synchronization service was called every 6 hours to handle failed synchronizations. The data collected during the collection period is merged into a signal table containing each sensor reading and their timestamp.

\subsubsection{Synchronization}
After storing the signals' table, the collector service will attempt to upload it directly to our web-server using wifi.
If wifi is not available on the smartwatch, the service will attempt to upload the data through the smartphone using the gateway device. 
This will create a proxy to the phone's Internet interface through the Bluetooth low-energy connection between the smartwatch and the smartphone. 

If both paths fail, the collection service stores the data in a temporary file on the smartwatch and terminates. When the synchronization service is activated, the delivery of all the previous failed synchronization files will be reattempted.

\subsection{Data processing pipeline}
In order to support the processing of the incoming data traffic from the smartwatches, we design a web-service, which contains different sections to handle requests' reception and verification, temporary hot storage, processing queue, storage queue, and a permanent storage. Our implementation is based on Python with Flask framework on top of an Apache web-server.
The requests contain one session of collected signals and the unique identifier that encrypts the username and password of the user that sent the data.
At the verification stage, the request credential token is verified against our users database. Any request that is not from an authorized user is dropped.

In order to increase the throughput of the web-server, the best practice is to close the connection as soon as it is not needed anymore. 
So after the verification stage the server will store the data in a temporary hot storage in the form of files and the address of the files and the associated username will be placed in the processing queue and storage queue.
At this stage, the web-server can close the connection and continue to process the other incoming requests.

We designed dedicated workers to process the data, described in detail in the next section. This process may result in a request of a new label for the data. If the system needs to collect a new label, the worker calls the notification sender API to send a push notification to the participants smartphone. Alongside the processing workers the system includes storage workers, which are given a lower priority compared to processing workers as they write the data into a permanent database. In our study, we use MongoDB. Our design is based on two main objectives: the optimization of the read/write time, and to ensuring scalable operations.

In order to optimize the operations, we first optimize the document size, as it directly affects the number of documents, number of operations per time interval, and the depth of the index tree in MongoDB database. We use a method suggested by MongoDB for time-series data called bucketing, where the data is chopped into buckets of maximum document size and each bucket is stored together as a single document. This adds another layer of folding and unfolding to the data but helps with the scale of the storage.

To address storage scale, we limit the data in both the hot and cold storage considering drive capacity limit, size of the index tree, and latency of queries. 

\subsection{Label collection}
Label collection takes the form of a questionnaire pushed to the user's smartphone by means of a smartphone app.
The mobile application is developed by Apache Cordova for both Android and iOS. The labeling request -- in our implementation triggered by a smart agent -- is processed by the notification service which reaches the user through the mobile app.

The questionnaire begins with a mood assessment, where the participants can choose between three main emotions: negative, neutral, and positive. The negative emotions then can be categorized in angry and sad. The participant then will be able to report the extent of each emotions by selecting one among three levels.  In the following page, the participants query specifically on their stress level (1-5).
In the final page of the questionnaire, the participants are asked to report their activity: first the physical state (sitting, standing, lying down, etc.), and then the specific activity (working, watching, reading, etc.).

\subsection{Experiments and Analysis}

One of the key metrics is latency. In the specific setting we address, a relevant latency measure is the time between data acquisition on the smartwatch and the time at which the corresponding label (if requested) is collected.
This time lapse can be divided into different components, 
some of which are determined by the environment and system architecture (e.g., queue time, learning time, upload time, notification delivery time), and some of which are determined by the participants (time to access the notification and fill out the questionnaire).

As the only period that is under our control is the total time to execute the processing pipeline, we focus our attention on that metric defined as the time between data submission to the web-server to the availability of the analysis result.
Notably, this time is a function of the number of users, which influences the load on the system queues.

\begin{figure}
    \centering
    \includegraphics[width=\linewidth]{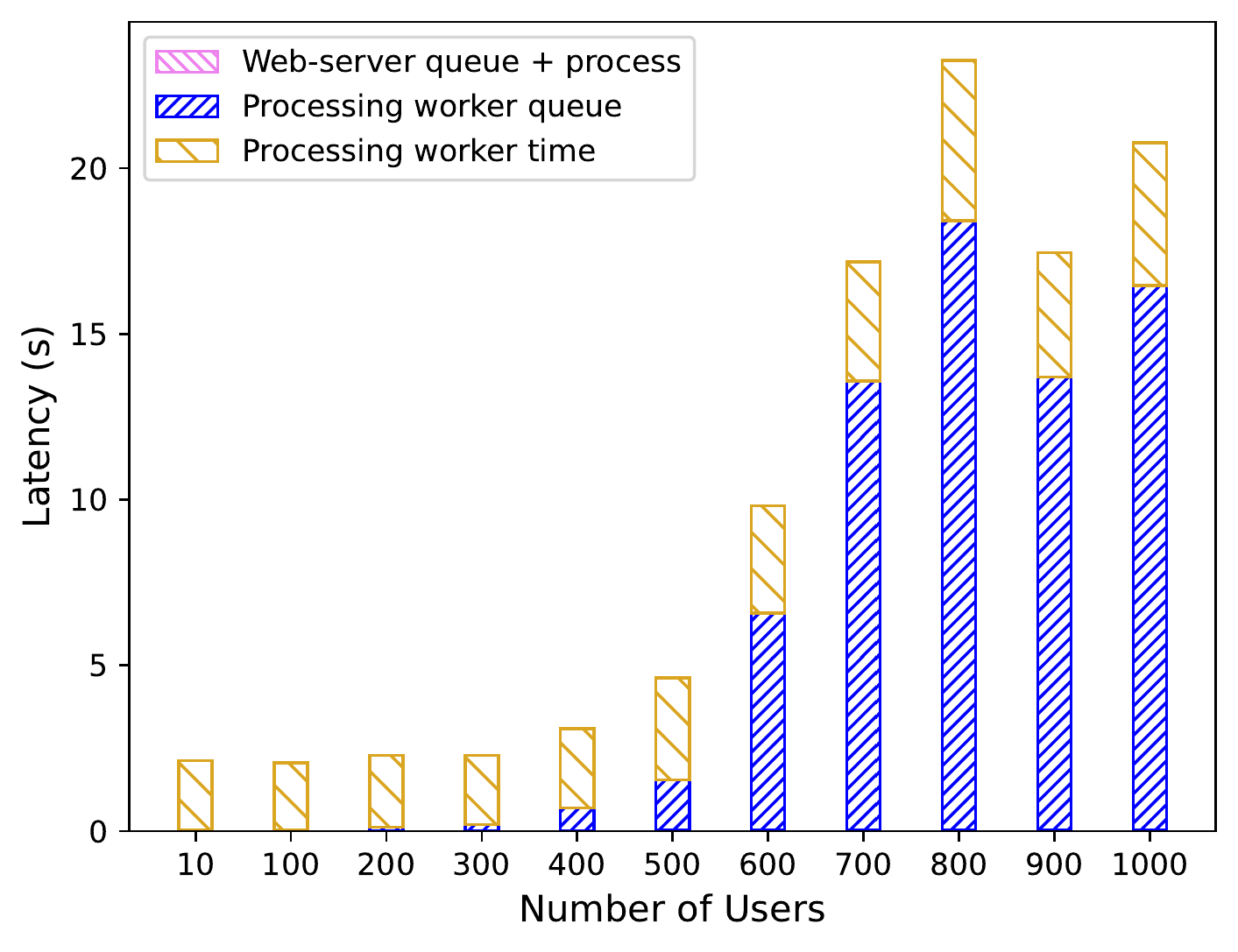}
    \caption{Average processing pipeline time as a function of the number of users.}
    \label{fig: latency}
\end{figure}

As shown in Fig. \ref{fig: latency}, the web-server latency is negligible compared to the processing time and processing queue time. This is due to the design of the web-server, which performs minimal functions and then closes the connection with the participants smartwatch or gateway device.
However, if the load is excessive, the processing queues will be filled with requests, which will eventually increase delay.
In our experiments, we observe that with less than 300 users in the system, the system can function with acceptable delay.
A perceivable delay increase can be observed if the number of users is in the range 300 to 500.
If the number of users exceeds 500, the queuing time is larger than the processing time, and the incoming data accumulate in the queues.

The experiments in this section have been ran on a simulated deployment of the processing pipeline over 3 VMs (virtual machines): one used for the web-server, one for the database, and one for the workers. This approach avoids issues such as memory race. Each virtual machine is using 2 cores of CPU and 2GB of RAM.
The measurements were performed using Apache JMeter\cite{JMeter}, an open-source performance testing tool in Java that can test the performance of web services.

\section{Dataset}
\label{sec: dataset}

We performed a data collection campaign on 20 volunteer subjects between June 2020 and June 2021. Subjects are selected in the population of undergraduate and graduate students. The population composition includes 13 males and 7 females, aged between 19-29 years. We collected a total of 109,586 samples throughout the study, over 2,629 days. Each subject participated in the study for a duration between 11 to 287 days with an average of 131 days, producing an average number of samples equal to 5479 instances from each subject. 

\subsection{Biosignals} As described earlier, the samples contain raw PPG, and motion signals including acceleration, gyroscope, and gravity.
The features extracted from the signals
are used for stress detection and for query decision making. The specific features we consider are: BPM, IBI, SDNN, SDSD, RMSSD, pNN20, pNN50, MAD,
SD1, SD2, S, SD1/SD2, and BR\footnote{
\textbf{BPM:} Beats per Minute, Heart Rate.
\textbf{IBI:} Inter-Beat Interval, average time interval between two successive heart beats (called NN intervals).
\textbf{SDNN:} Standard Deviation of NN intervals.
\textbf{SDSD:} Standard Deviation of Successive Differences between adjacent NNs.
\textbf{RMSD:} Root Mean Square of Successive Differences between the adjacent NNs.
\textbf{pNN20:} The proportion of successive NNs greater than 20ms (or 50ms for pNN50).
\textbf{MAD:} Median Absolute Deviation of NN intervals.
\textbf{SD1 and SD2:} Standard Deviations of the corresponding Poincaré plot.
\textbf{S:} Area of ellipse described by SD1 and SD2.
\textbf{BR:} Breathing Rate.}.

\subsection{Labels} 
Among all participants, a total of 3411 labels with the answer to at least one question are collected. The distribution of the collected labels is presented in Table~\ref{tab: labels}. From the self reported labels \textit{a little bit}, \textit{neutral}, \textit{sitting}, and \textit{working with computer} were the most common answers to the questions in order, which might be biased due to the fact that the participants were mostly university students and scholar.

\begin{table}
    \caption{Distribution of collected labels}
    \begin{tabular}{|p{1.2cm}|p{1.2cm}|c|c|p{1cm}|p{1cm}|}
    \multicolumn{6}{c}{\textbf{Stress}}\\ \hline
    not at all          & a little bit       & some     & a lot   & \multicolumn{2}{c|}{extremely} \\ \hline
    625                 & 1636               & 956      & 184     & \multicolumn{2}{c|}{10} \\ \hline
    \multicolumn{6}{c}{\textbf{Emotions}} \\ \hline
    happy               & \multicolumn{2}{c|}{neutral} & \multicolumn{2}{c|}{sad} & mad \\ \hline
    1188                & \multicolumn{2}{c|}{2061} & \multicolumn{2}{c|}{80} & 59  \\ \hline
    \multicolumn{6}{c}{\textbf{Physical Activity}}\\ \hline
    lying down          & sitting            & standing & walking & jogging     & working out \\ \hline
    391                 & 1675               & 107      & 159     & 15          & 22\\ \hline
    \multicolumn{6}{c}{\textbf{Main activity in the past 5 minutes}}\\ \hline
    {\scriptsize working with Computer} & {\scriptsize working with smart phone} & speaking & reading & watching TV & resting\\ \hline
    1095                & 455                & 309      & 38      & 296         & 247 \\ \hline
    \end{tabular}
    \label{tab: labels}
\end{table}

\section{Stress Detection based on bio-signals}
We tested a number of classifications methods to detect stress based on PPG features including: Support Vector Machine (SVM), XGBoost (XGB), and Random Forest (RF). 
Support Vector Machine is a classification (and regression) method that maximizes the distance of a hyperplane from the nearest points from each of the two classes. We used SVM with a radian basis function.
XGBoost is a faster and more accurate version of gradient boosted decision trees, which also has built-in regularization that reduces overfitting.
Random Forest is an ensemble learning method that consists of a `forest' of decision trees. Each decision tree is trained on a subset of features of a subset of the training data (sampled with replacement), and RF takes the `mode' (classification) of the output of the individual trees as the predicted class. Hyper-parameters such as number of trees, maximum depth of a tree, and the minimum number of samples to split a node inside a tree are optimized using grid search.

In the collected data, the stress labels are in five levels, we map them to binary classes: stressed vs not-stressed. We train and test each classifier using a leave-subject-out approach, meaning the training is performed on data from all subjects except one, and then the model is tested on the data from the left-out subject. We then average on all subjects and report the result. Given the particular application of stress detection in everyday-settings, we focus on recall (True Positive Rate), and the the result of the analysis is in Table~\ref{tab: classification}. Each row corresponds to a different mapping, for instance in the last row the first low stress labels are mapped to `not stressed' and the two high stress labels are mapped to `stressed'. The Random Forest performs slightly better than the other two models as the system acquires more labels. We then select RF as classification method in the rest of this paper.

\begin{table}[bt]
    \centering
    \vspace{.15cm}
    \caption{Average Recall (on class \textit{stressed}) on the two classes, using leave-subject-out Cross-Validation method. Stress labels are five levels: \textit{not at all} (0), \textit{a little bit} (1), \textit{some} (2), \textit{a lot} (3), \textit{extremely} (4) which are mapped to binary classes.}
    \scalebox{1}{
    \begin{tabular}{l l l l l l} 
    \specialrule{0.1em}{0.1em}{.1em} \\[-3ex]
    &  &  & \multicolumn{3}{c}{Leave-subject-out}\\
    &  &  & \multicolumn{3}{c}{Cross-Validation (Recall)}\\
    \cline{4-6}\\[-5ex]
    not & & No.~of & & & \\[-1ex]
    stressed & stressed & instances & SVM & XGB & RF\\[1ex]
    \hline \\[-2ex]
    0       & 1     & (557, 1286)   & 0.09 & 0.31 & 0.29 \\
    0       & 2     & (557, 822)    & 0.11 & 0.36 & 0.31 \\
    0       & 3, 4  & (557, 154)    & 0.15 & 0.34 & 0.41 \\
    0, 1    & 3, 4  & (1843, 154)   & 0.21 & 0.31 & 0.37 \\
    0, 1, 2 & 3, 4  & (2665, 154)   & 0.20 & 0.33 & 0.36 \\ [1ex]
    \hline\\ [-4ex]
    \end{tabular}}
   \label{tab: classification}
\end{table}

\section{Context Aware Active Learning}
\label{sec: caal_method}

Supervised learning algorithms typically use all the available labeled data to train a model. As explained earlier, in our personalization setting we have access to labels previously collected from other individuals, while labels from the currently monitored subject need to be collected by means of queries. Then the problem of determining how to request a query arises. In offline active learning, an algorithm chooses which instances in the overall dataset are used as training data. The problem is often structured in a sequential fashion, where a learner is first trained on a small set of available labeled data, and then the algorithm selects one by one or in groups the most informative unlabeled instances and query for labels from an oracle (e.g. a human annotator). 

AL algorithms generally aim at obtaining as much performance gain as possible by using as few labeled instances as possible. Equivalently, they select the most useful instances in the dataset and and then try to get a label for them, and they discard useless or misleading instances. In many circumstances the labels are provided by an oracle (e.g. a human), and a cost is associated with each provided label, so that AL methods help to reduce this cost as much as possible while maintaining the performance or even improving it. Related research in AL is quite rich and explores the design of query rules. The principal query strategies include the uncertainty based approach \cite{seung1992query}, diversity based approach \cite{nguyen2004active}, and expected model change based approach \cite{freytag2014selecting, roy2001toward}. 

From a practical perspective, AL approaches can be divided into two general categories: stream-based selective sampling, and pool-based sampling ~\cite{krishnamurthy2002algorithms, mccallumzy1998employing}. The key difference between the two is that the former makes a judgement on whether each instance in the stream needs to query the labels independent from others, while the latter makes the judgement based on the entire pool of data and selects the best N instances.
In traditional AL it is commonly assumed all instances are equally available to be labeled at the same cost. There are some expanded forms of active learning in which different instances require different costs to be labeled \cite{settles2008active, song2017contextual}.

The problem at hand is significantly different. In fact, while we can generally associate the user effort in responding to a label request as a cost, so that we attempt to minimize effort while maximizing detection performance, there is a considerable impact of user behavior factors on the quality and availability of labels. In fact, the user will to respond, as well as the timing of response, may depend on a number of contextual variables, such as the current activity or the time of the day. While a failure to respond prevents access to the label, a delayed response may decrease the correlation with the target sample -- thus degrading the quality of the dataset -- or may lead to associate the label to another sample (e.g., if the closest sample is paired with labels). Moreover, there likely is a temporal correlation effect, by which the user may be unwilling to respond to a request if he/she just responded to a previous query. We note that temporally neighboring labels may refer to highly correlated, and then less useful, samples.

Thus, we contend that in our setting, active learning needs to account for the context determining the user response, as well as for the implications on future labels availability if a request is issued. For these reasons, we propose to adopt an approach based on Deep Q-Learning, where the agent incorporates in the state and reward function the notion of context.

\subsection{Deep Q-Learning}

We provide a brief overview of Q-Learning and Deep Q-Learning, and we refer the interested reader to one of the many resources available in the literature that discuss in-depth this class of algorithms. The objective of the algorithm is to select an action $a$ based on the current state $s$ of the overall system. 
The core driver of the algorithm is a function that measures the \textit{quality} (Q-value) of a state-action combination:
\begin{equation}
    Q : S\times A \rightarrow \mathbb{R}.
\end{equation}
Q-values represent the expected long-term reward (or cost) of taking a certain action given the current state. In traditional Q learning, a table of these Q-values mapped on the $(s, a)$ pair is initialized and then updated based on a predefined rule as the agent interacts with the system. The most popular form of this rule is the Bellman equation which updates the Q-values iteratively as follows:
\begin{equation}
    \label{eqn: bellman}
    Q^{new}(s_t, a_t) = (1-\alpha)Q(s_t, a_t) + \gamma (R_t + \alpha \maxA_a Q(s_{t+1}, a)),
\end{equation}
where $t$ is a temporal index, $R_t$ is the reward at time $t$,  $\alpha$ is the learning rate which controls the convergence speed through scaling the impact of new incoming information vs previously observed information on the final Q function. $\gamma \in [0, 1]$ is the discount factor which controls the importance of future rewards; a small $\gamma$ trains a \textit{myopic} agent which only considers current or short term rewards, while a larger $\gamma$ trains the agent to strive for long-term high rewards. The policy is then set to select the action with the highest Q-value in each state $s_t$ (in the exploitation phase post training):
\begin{equation}
    \label{eqn: policy}
    a_t = \argmaxA_a Q(s_t, a). 
\end{equation}
in Deep Q-Learning, instead of a table of Q-values, a neural network is utilized to approximate the Q-values. In the case where the actions are a limited number of discrete values (e.g. binary actions), a common approach is to implement the Q network in the form of a fully connected neural network with N (number of state features) input nodes and M output nodes (number of possible actions), and each output node provides the estimated Q-value for the corresponding action. Then the Bellman equation (Eq.~\ref{eqn: bellman}) is used to train the neural network and optimize the network weights.

In the following, we define the state and action space, as well as the reward function and network architecture, we used in our implementation.

\subsection{State Space}
\label{subsec: state_space}
The state variable $\mathbf{s}$ is a vector that captures the information that results in a certain action to be taken. Our ultimate goal is to train and personalize the stress detector to achieve high accuracy while submitting a small number of queries to the user. An existing pre-trained classifier has gives a confidence level for each prediction. We remark again that collecting labels for the instances that are predictable with high confidence is unnecessary and inefficient, because using ``obvious'' data points to train of the classifier is not expected to improve the performance significantly. So confidence level of the existing classifier on the instance is one factor in filtering instances for query. Confidence level measures how close the raw output of the classifier is to the decision boundary, which is directly related to the raw output of the pre-trained classifier, in this paper we use this raw output as a representative of confidence score, and refer to it as uncertainty factor. The uncertainty factor is usually the only criteria that is used in active learning algorithms to filter instances. In our setting, the system can acquire features that are correlated with the state of the user to make the agent aware of the context, and its correlation with relevant factors such as response rate, response time, etc.  As an example, users are probably more responsive in certain hours during daytime compared to night, or they are probably more responsive when they are not working. As a representative of these contextual information, we use time of day (split in one hour intervals) and learn a response rate online for each time interval. We then use these response rates as state variable in the Q-network.

In addition, in order to improve user experience by submitting too many queries in a short time span, as well as to encourage a representative distribution of labels we include in the state the time to the last query.

The complete list of the state variables that constitutes the overall state vector is presented in Table~\ref{tab: state_variables}. The range column shows each variable's range pre-normalization, these variables are then normalized (to the interval $[0, 1]$ before they are used as a network input. For the \textit{time distance} variable, we clip it at the 180 minutes mark, and then map the values linearly to the unit interval.
\begin{table}[tbp]
    \caption{components of the state space}
    \begin{center}
        \begin{tabular}{|l|l|} 
             \hline
             \textbf{Factor} & \textbf{Range}\\ \hline
             Uncertainty region & [0, 1] \\
             Time distance to the last query & $\{15, 30, 45, ...\}$ minutes\\
             Response rate & [0, 1] \\
             Time of day & $\{0, 1, 2, ..., 24\}$\\
            \hline
        \end{tabular}
        \label{tab: state_variables}
    \end{center}
\end{table}

For each new instance that is received at the cloud, after preprocessing and feature extraction, the stress detector predicts whether it corresponds to a stress event or not. The direct output of the classifier is a float number in the interval $[0,1]$ which then is then mapped to $\{0,1\}$, but this float number also gives us a measure of the classifier confidence on the instance. For instance, assuming the decision boundary is $\theta=0.5$, the closer the raw output is to this point the lower the confidence, and vice versa. Accordingly, other state variables are calculated, and we use them to build the state vector. This state vector is then used as the input for the trained Q-network to determine the optimal action. 

\subsection{Reward Function}
During the training phase, Eq.~\ref{eqn: bellman}, for each instance we need to assign a certain reward for all the possible actions that can be taken. In the proposed model, the reward function is designed such that instances that are desirable to be queried for labels are rewarded a higher value for for action 1 (a low reward for action 0), and vice versa. The criteria to define the reward function must be adjusted to capture the desirable properties for both actions. Based on the discussion in Section~\ref{subsec: state_space}, we select the uncertainty factor of the classifier, time distance to the previous query, response rate in the context of the instance as driving metrics to design the reward function. We use the sigmoid functions with two parameters to select desirable and non-desirable regions of each variable. The sigmoid function is defined as:
\begin{equation}
    S(x; \alpha, \beta) = \frac{1}{1+e^{-\alpha(x-\beta)}},
\end{equation}
where $\beta$ is the center of transition, and $\alpha$ is the transition scale. If we chose a large $\alpha$, the function collapses into a step function that is 0 before $\beta$ and 1 after that. For smaller values of $\alpha$ the transition becomes smoother. If we consider each state variable as $\mathbf{s}=(s_1, ..., s_4)$, then the respective components of the proposed reward function are:
\begin{equation}
    \label{eqn: reward_comps}
    r_1(s) = S(s_1; 100, 0.5) \times S(s_1; -100, 1.3),
\end{equation}
\begin{equation*}
    r_2(s) = S(s_2; 10, 0.3), \quad 
\end{equation*}
\begin{equation*}
    r_3(s) = S(s_3; 100, 0.4)),
\end{equation*}
and the proposed reward function is:
\begin{equation}
    \label{eqn: reward_func}
    R(s, a=1) = 2 \cdot r_1 \cdot r_2 \cdot r_3,
\end{equation}
\begin{equation*}
    R(s, a=0) = 1-r_1 \cdot r_2 \cdot r_3.
\end{equation*}
In these equations $a=1$ represents action \textit{submit query}, and $a=0$ represents action \textit{do not submit query}. The values of parameters $\alpha_i, \beta_i$ in Eq.~\ref{eqn: reward_comps} are adjusted such that the reward component $r_i$ has a value close to 1 for the desirable regions, and a value close to 0 for non-desirable regions. Since the number of instances for which $a=1$ should be taken is significantly lower than other instances, the coefficient 2 helps with a faster and more stable convergence during training of the agent.

\begin{figure}
    \centering
    \includegraphics[width=1\linewidth]{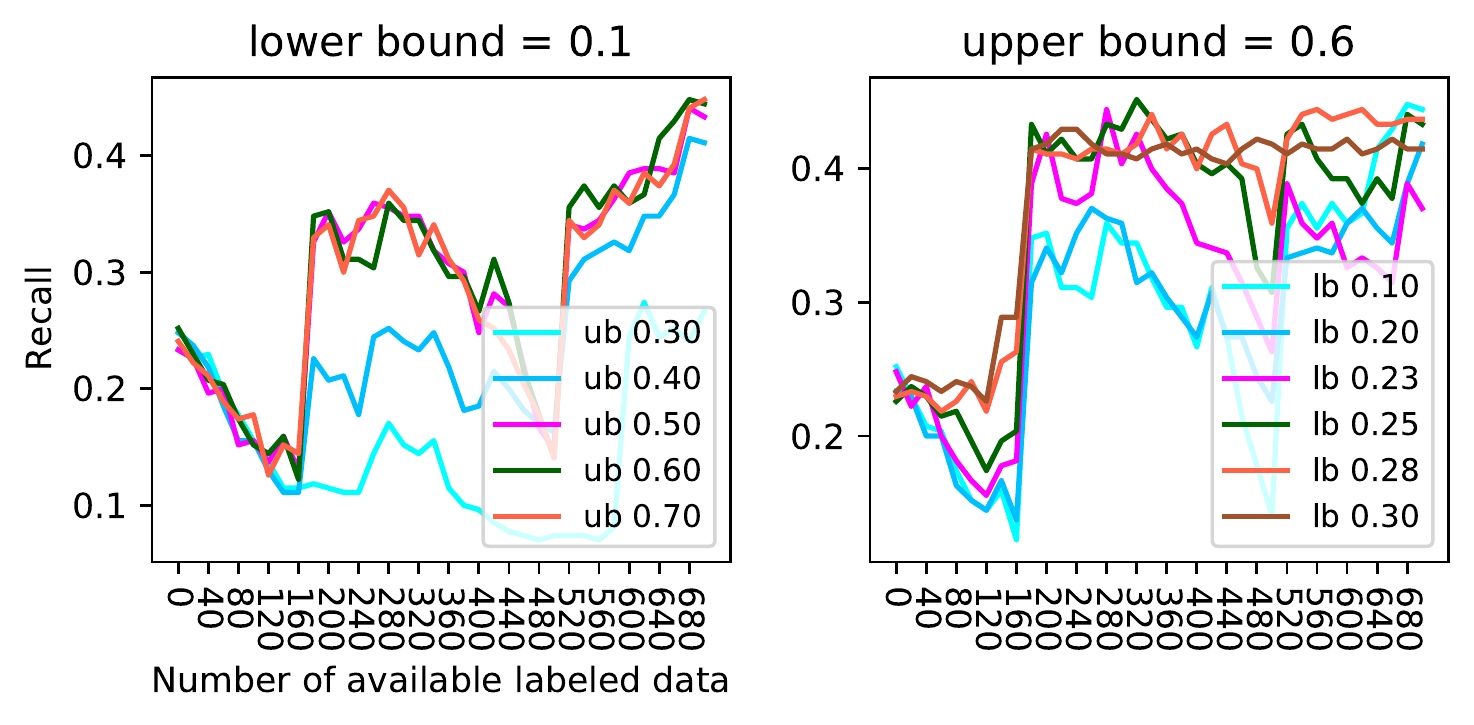}
    \caption{Effect of lower bound and upper bound of the classifier output region used for active learning.}
    \label{fig: boundaries}
\end{figure}

 We performed an analysis to find the practical region of the raw output to select or reject instances. First, we keep lower bound fixed at $t_1=0.1$ and inspect the personalization process when the upper bound is set at a few different points. It is clear in Fig.~\ref{fig: boundaries} (left) that personalization improves when the upper bound shifts up to the point $t_2=0.5$, but not much after that (even though the number of instances that pass the filter increase with $t_2$). We run a similar analysis for the lower bound, by keeping the upper bound fixed at $t_2=0.6$, and as the outcome in Fig.~\ref{fig: boundaries} (right) suggests, $t_1<=0.25$ are almost identical. We performed this analysis on data from multiple subjects, which shows the region boundaries might be slightly different, but they are always around the mid interval of $[0,1]$. Therefore, to be inclusive of the optimal regions for different subjects, we set the region $[t_1, t_2]$=$[0.2,0.6]$ as the criterion for active learning, in the rest of this paper.

\subsection{Q-network and Policy function}
\begin{figure}[tbp]
\centerline{\includegraphics[width=\linewidth]{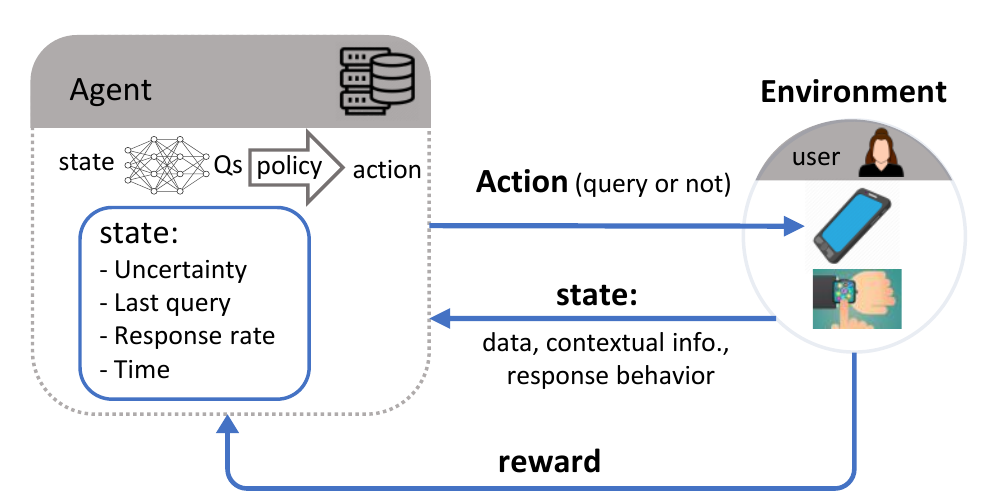}}
\caption{Architecture of the proposed Deep Q-Network.\vspace{-2mm}}
\label{fig: dqn}
\end{figure}

The overall architecture of the proposed model is shown in Fig.~\ref{fig: dqn}. We use a fully connected neural network for the estimation of the Q-values. We achieve the best performance using a network with two hidden layers, and 128 nodes in each layer. The network has 4 input nodes (number of state variables), and 2 output nodes (number of possible actions).
Bias terms are added and and both l1 and l2 kernel regularization terms are added to every hidden node. The policy function $\pi_{\theta}(s, a)$ selects the action that results in a higher Q-value for the given state $\mathbf{s}$. In addition, for small portion of instances ($\sim5\%$) a query is randomly issued to count for potential unobserved sections of the state space (exploration).

After every K=100 instances (approximately once every 24 hours in ideal conditions), the response rate values for the user is updated based on their response behavior, this counts for personal differences and updates the query selection strategy for each subject.
In addition, the stress detector is retrained using the new subjective labeled data added to the the old objective (from other subjects) data. 

\section{Results and Discussion}
\label{sec: results}
We start by pre-training the stress detector, which will also be used to extract the state and reward in the construction of the active learning framework. We use a random forest classifier with $n=500$ estimators (number of trees) and $max\_depth=5$ for each tree. We map the stress labels into two classes: 
\textit{a lot} and \textit{extremely} as $y=1$ (stressed), and the other three labels as $y=0$ (not stressed). We train the model on labeled data from 14 different subjects (leave one subject out for personalization). The pre-trained model on the new subject gives a recall value of $recall=0.238$ for the minority class (class \textit{stressed}).

The Q-learning agent is pre-trained on an offline sequential objective dataset, and when we reach the end of the sequence (called one episode), we start over from the beginning until we reach the total number of steps. We trained the model for K=200,000 steps. The total reward achieved on the episodes saturates before this point, which suggests the model has converged. 
%
%

The agent is trained in two different modes. First, we use a ``traditional'' formulation for the agent where we only focus state and reward on the classifier's raw output. The agent is highly rewarded to take action `submit query' for the instances in the uncertainty region of the classifier, and rewarded to take action `do not submit query' for other instances. In this mode, the agent does not capture the contextual information and acts as a traditional active learning selection policy.
Then, we include the contextual information in the reward function as described earlier: we highly reward action `submit query' for instances that not only are in the uncertainty region, but also are from a time interval which the user has shown to be more responsive around that time, and the instances are not in a short time distance from the previous action `select'. For other instances the agent is rewarded to take action `do not submit query'.

\begin{figure*}[htbp]
     \centering
     \begin{subfigure}[b]{\textwidth}
         \centering
         \includegraphics[width =\textwidth]{ 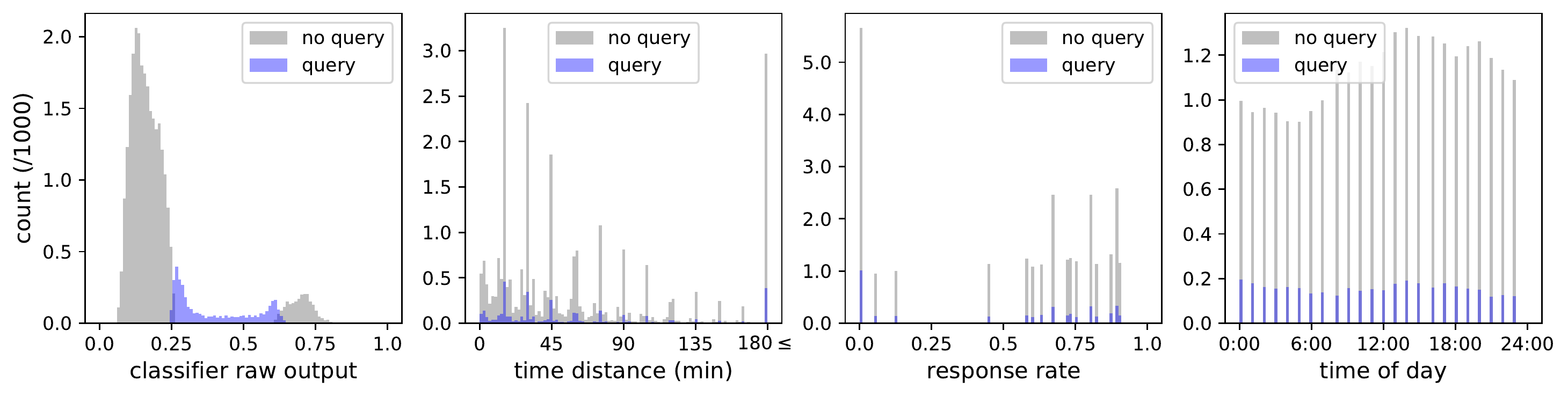}
         \caption{Non-contextual active learning\vspace{-2mm}}
         \label{fig: al_agent}
     \end{subfigure}
     \par\bigskip
     \begin{subfigure}[b]{\textwidth}
         \centering
         \includegraphics[width =\textwidth]{ 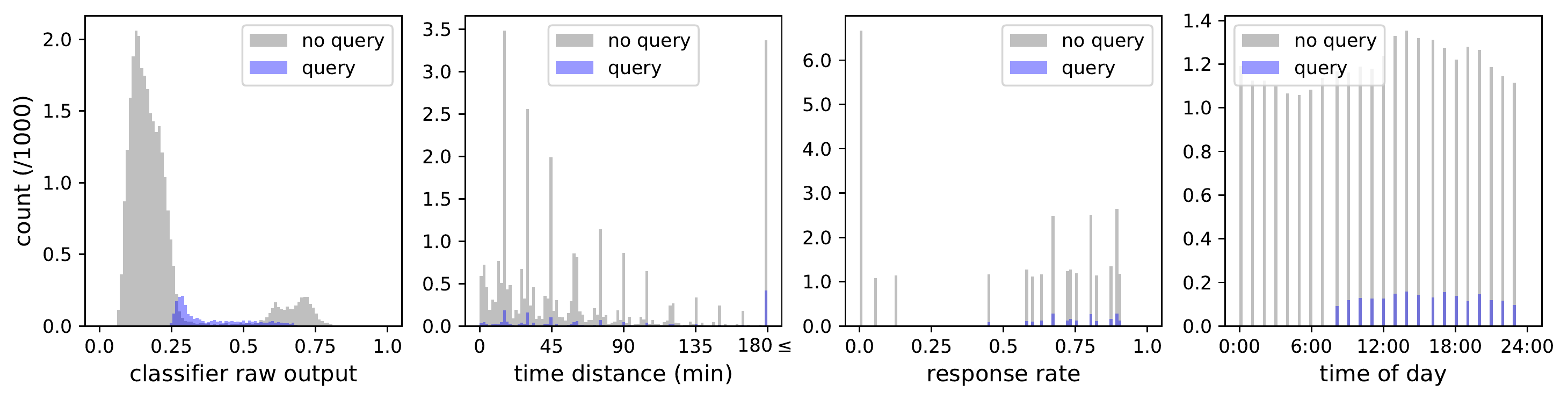}
         \caption{Context aware active learning}
         \label{fig: caal_agent}
     \end{subfigure}
        \caption{After training, distribution of the state variables for the instances that are selected to be labeled (blue), and the instances that are not selected (gray). Vertical axis are the number count.\vspace{-2mm}}\vspace{-2mm}
        \label{fig: states_dist}
\end{figure*}

The distribution of state variables of instances for which action `submit query' and `do not submit query' is taken using each of the two agents is shown in Fig.~\ref{fig: states_dist}. In Fig.~\ref{fig: al_agent}, the first agent considers \textit{being from the uncertainty region of the classifier} as the \textit{necessary and sufficient} condition for the instance to be selected. However, the second agent uses the uncertainty region only as a \textit{necessary} condition to select the instances, as presented in Fig.~\ref{fig: caal_agent}.
As shown in the left-most plot, not every instance that meets the criteria of uncertainty region is assigned the action \textit{select}. The instances that meet the uncertainty criteria but are not in the `select' distribution are three main groups: first instances that are close in time to the previous `select' instance, to avoid successive queries - for better user experience, and also the fact that instances that are close in time probably share similar stress labels.
The second group are instances that are from a time interval in which the user is not responsive (sleeping, working, away from their phone, etc.). And the third group are instances that are not from the main uncertainty region or from a time of high response-rate, and the agent decides to \textit{wait} for the following instances that it predicts to be more impactful instances, and there is a higher chance to get the response for the query. 

\begin{figure}[htbp]
\centerline{\includegraphics[width=0.5\textwidth]{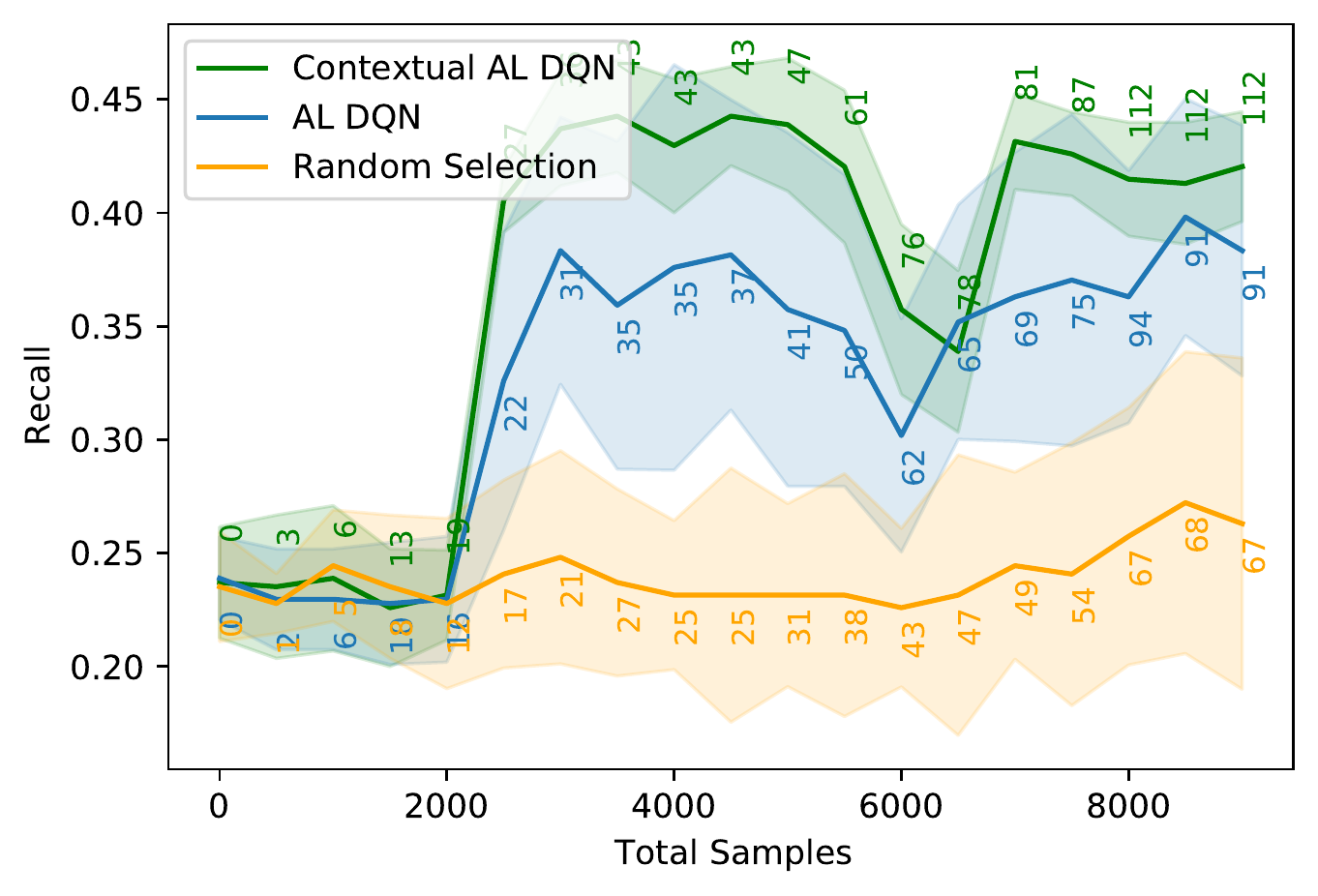}}
\caption{Personalizing the model using different instance selection methods.}
\label{fig: personalization}
\end{figure}

We evaluate the performance of each of these two agents as well as a random selection policy on the ultimate goal -- personalization of the classifier -- and the results are depicted in Fig.~\ref{fig: personalization}. The number of selected instances (for query) with each selection method at each step is the same. However, a portion of the selected instances remain unlabeled, and therefore the number of instances that are available for personalization is different when we utilize different selection policies. Other than that, the selected instances with different policies have different impact on personalization (random selection vs. the other two). We have a total of 12,700 instances from this one subject, which 922 of them are labeled. We kept 230 labeled instances (from the end of the sequence) from one subject as test data, and the rest are available for training ($25\%-75\%$). We start by no personalization, and gradually move forward in the partially labeled subjective data. A portion of this data is selected for query, and a portion the selected data is labeled, the labeled data is used for personalization. At each step the number of selected data by the context-aware agent is lower than the number of selected data by the non-context aware agent, so we had to down-sample the number of queries from the larger group, to provide a fair comparison between the agents, we did this down-sampling randomly. Also for random selection, we randomly selected a number of partially labeled samples equal to the number of queries from the agents. To reduce the effects of random selection, at each step we selected the instances, and personalized the models N = 100 times. For the three selection methods, the mean and standard deviation of recall (True Positive Rate) for the class \textit{stressed} on test data are presented in Fig.~\ref{fig: personalization}.

Instances that are selected by the random agent do not improve the performance significantly since they include samples from the entire region of the input space of the classifier, including samples whose class is `trivial' to be extracted. Instances that are selected by a non-contextual active learning method (blue curve) increase the performance. However, with an equal number of queries, the best result is achieved when the agent is context aware (green curve), since it results in a higher number of impactful instances which also have a higher chance to receive the label.

\begin{figure}
    \centering
    \includegraphics[width=\linewidth]{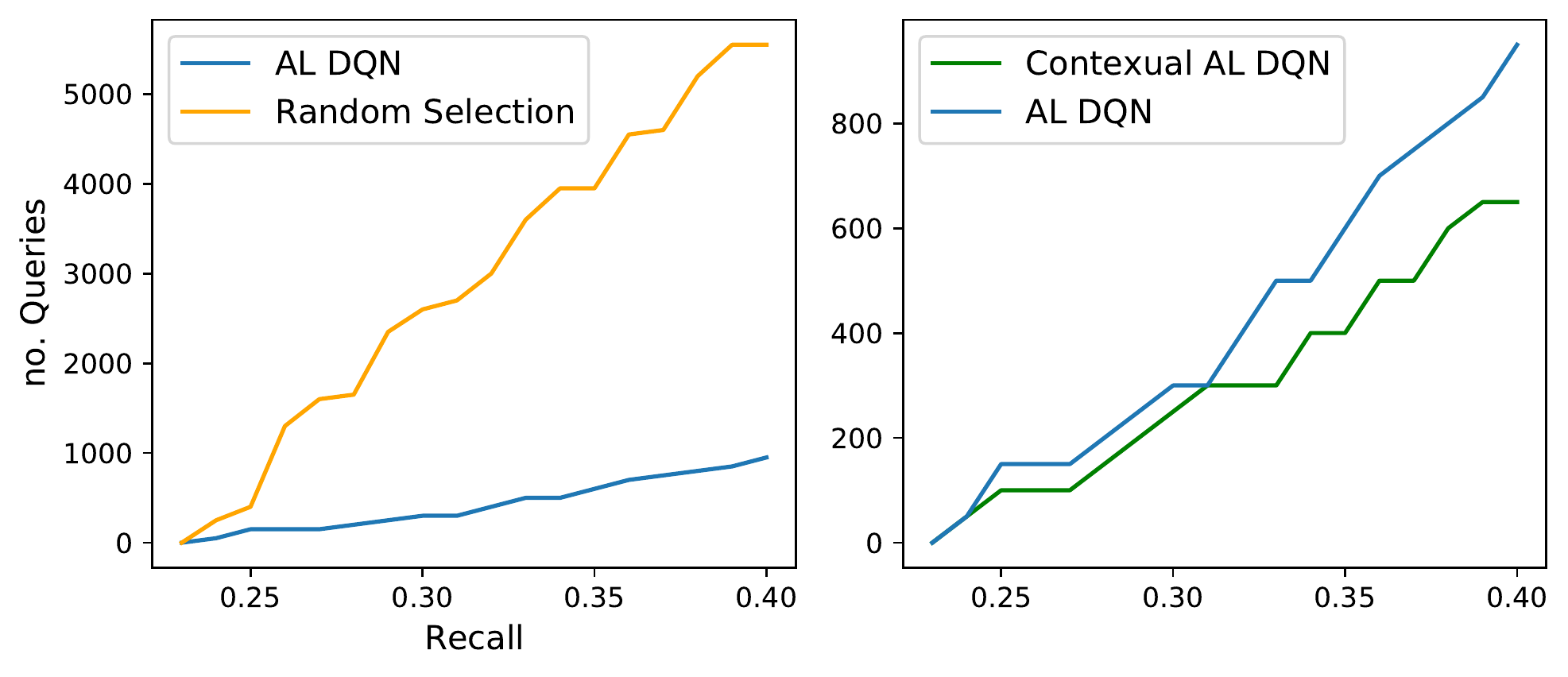}
    \caption{Number of queries needed to reach a certain performance level during personalization.}
    \label{fig: no.queries.labels}
\end{figure}

We also analyze the number of queries required to reach a certain level of personalization. We repeated the experiment N = 100 times to reduce the impact of random selection. The average number of instances is shown in Fig.~\ref{fig: no.queries.labels}. There is a wide gap between random selection versus the DQN agents, and the context-aware agent makes it possible to reach high performance with a lower number of queries. Specifically, the context-aware selection policy, reduces the number of required queries up to $88\%$ compared to random selection method. Between the two DQN agents even though the number of required \textit{labels} remains identical throughout the analysis (which is expected), the number of required queries is reduced by up to $32\%$ in the case of the context-aware agent. These results are based on data from the subject from whom we had a higher number of labels, but we observed similar trends on data from other subjects.

\begin{figure}
    \centering
    \includegraphics[width=\linewidth]{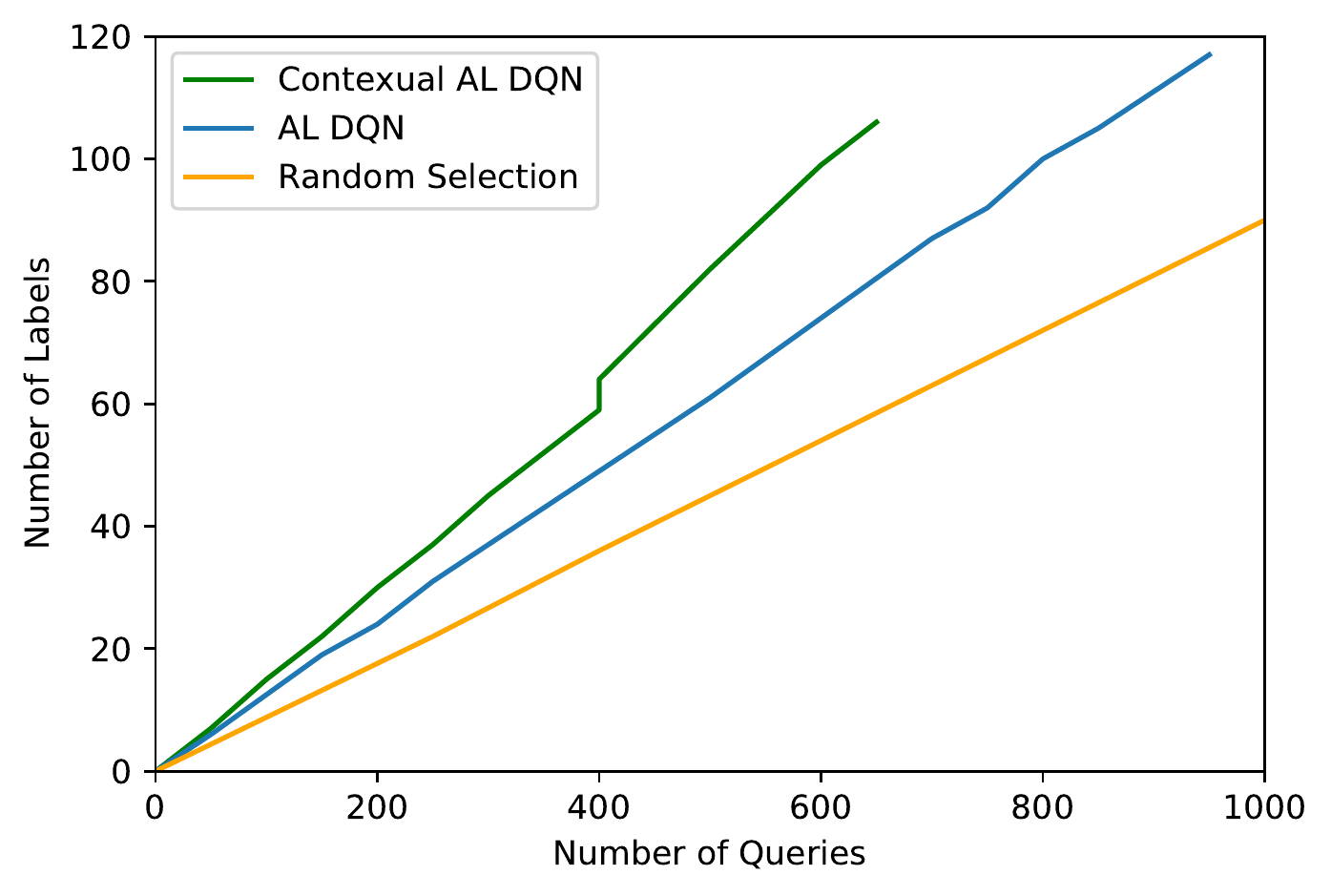}
    \caption{Response rate behavior for different query methods}
    \label{fig: response_rate}
\end{figure}

In addition, Fig.~\ref{fig: response_rate} depicts the response-rate patterns associated to different query methods. The contextual agent improves the response rate as expected, but also the non-contextual agent improves the response rate. This latter effect is due to a certain amount of correlation between unresponsive states of the user (e.g., sleep) and the uncertainty of the classifier.

\section{Conclusions}
\label{sec: conclusion}
Detecting physiological stress and other indicators of mental well-being based on physiological data in everyday settings is a challenging task, mainly due to bias in both signal and labels. Personalizing the detection models for each subject significantly improves prediction performance. We developed a multilayered platform to transfer and analyze biosignals in real-time and control online data collection. To optimize the interaction with the user and facilitate personalization, we developed a Q-learning based active learning agent, which is able to not only identify impactful samples, but also capture subjects' response patterns based on the context. 

We collected a dataset of physiological data along with well-being and activity labels in a long time-frame in everyday settings, which we pledge to release. The context-aware active learning agent we proposed reduces the number of required queries to reach the desired level of personalization compared to random selection, and a non-context aware agent, by up $88\%$ and $32\%$ respectively. 


\begin{acks}
This work was partially supported by NSF Smart and Connected Communities (S\&CC) grant CNS-1831918. 
\end{acks}

\bibliographystyle{ACM-Reference-Format}
\bibliography{Bibliography}

\end{document}